\documentclass[11pt,a4paper]{article}

\usepackage[table,dvipsnames]{xcolor}
\usepackage{emnlp2020}
\usepackage{times}
\usepackage{calc}
\usepackage{amsmath}
\usepackage{latexsym}
\usepackage{booktabs}
\usepackage{tikz}
\usepackage{tikz-qtree}
\usetikzlibrary{decorations.text, decorations.pathreplacing}
\usepackage{url}
\usepackage{multirow,rotating}
\usepackage{booktabs,array,xcolor,makecell}
\usepackage{microtype}
\usepackage{hyphenat}
\usepackage[scaled=0.89]{helvet} 

\newcommand{\formal}[1]{\textsf{{#1}}}
\newcommand{\relation}[1]{\formal{\textit{#1}}}

\makeatletter
\newcommand*{\@rowstyle}{}
\newcommand*{\rowstyle}[1]{
  \gdef\@rowstyle{#1}%
  \@rowstyle\ignorespaces%
}
\newcolumntype{=}{
  >{\gdef\@rowstyle{}}%
}
\newcolumntype{+}{
  >{\@rowstyle}%
}
\makeatother

\aclfinalcopy 
\def\aclpaperid{1103}

\definecolor{bkgray}{gray}{0.8}
\title{GLUCOSE: GeneraLized and COntextualized Story Explanations}

\author{Nasrin Mostafazadeh\Thanks{Current affiliation Verneek, Inc.} \hspace{0.3in}
   Aditya Kalyanpur \hspace{0.3in}
   Lori Moon \hspace{0.3in} 
   {\bf David Buchanan\Thanks{Current affiliation QuillBot Inc.}} \\
   {\bf Lauren Berkowitz} \hspace{0.3in}
   {\bf Or Biran}
  \hspace{0.3in}
   {\bf Jennifer Chu-Carroll} \\
   Elemental Cognition \\
   New York, NY, USA \\
   {\tt nasrin@verneek.com}\\
  {\tt \{adityak, lorim, orb, jenniferc\}@elementalcognition.com}\\
  {\tt david.buchanan@quillbot.com}
   }

\date{}

\begin{document}
\maketitle
\begin{abstract}
When humans read or listen, they make implicit commonsense inferences that frame their understanding of what happened and why. As a step toward AI systems that can build similar mental models, we introduce GLUCOSE, a large-scale dataset of implicit commonsense causal knowledge, encoded as causal mini-theories about the world, each grounded in a narrative context. To construct GLUCOSE, we drew on cognitive psychology to identify ten dimensions of causal explanation, focusing on events, states,  motivations, and emotions. Each GLUCOSE entry includes a story-specific causal statement paired with an inference rule generalized from the statement. This paper details two concrete contributions. First, we present our platform for effectively crowdsourcing GLUCOSE data at scale, which uses semi-structured templates to elicit causal explanations. Using this platform, we collected a total of \~{}670K specific statements and general rules that capture implicit commonsense knowledge about everyday situations. Second, we show that existing knowledge resources and pretrained language models do not include or readily predict GLUCOSE's rich inferential content. However, when state-of-the-art neural models are trained on this knowledge, they can start to make commonsense inferences on unseen stories that match humans' mental models.
\end{abstract}

\section{Introduction}

\begin{table*}[tb]
    \newcommand{\sentence}[1]{
        \path [
            decoration={
                text=#1,
                text effects along path,
                text effects/.cd, 
                path from text, text along path,
                group letters, word count=\w,
                every word/.style={name=word-\w, execute at begin node=\strut, inner sep=0pt, outer sep=0pt}
            },
            decorate
        ] (0,0)
    }
    \newcommand{\spanannotation}[3]{
        \draw [decoration={brace, mirror, raise = -0.03cm}, decorate] 
          (word-#1.south west) -- (word-#2.south east)
          node [midway, inner sep=0pt, outer sep=0pt, yshift=-0.2cm] {\tiny #3\strut}
    }
    
    \small
    \centering
    \setlength{\tabcolsep}{3pt}
    \renewcommand{\arraystretch}{0.8}
    \begin{tabular}{m{2cm}l}
    \toprule
    \textbf{Dimension} & \textbf{Semi-structured Specific Statement and Inference Rule}:  antecedent {\scriptsize\relation{connective}} consequent \\ \midrule
    \multirow{3}{=}[0.5\baselineskip]{1: Event that directly causes or enables $X$}
    & \begin{tikzpicture}
        \sentence{A car turned in front of him {\scriptsize\relation{Causes/Enables}}  Gage turned his bike};
        \spanannotation{1}{2}{subject};
        \spanannotation{3}{3}{verb};
        \spanannotation{4}{6}{preposition};
        \spanannotation{7}{7}{object};
        \spanannotation{9}{9}{subject};
        \spanannotation{10}{10}{verb};
        \spanannotation{11}{12}{object};
    \end{tikzpicture}
   \\
     & \cellcolor{bkgray}\begin{tikzpicture}
        \sentence{Something$_A$ turns in front of Something$_B$ (that is Someone$_A$'s vehicle) {\scriptsize\relation{Causes/Enables}}};
        \spanannotation{1}{1}{subject};
        \spanannotation{2}{2}{verb};
        \spanannotation{3}{5}{preposition};
        \spanannotation{6}{10}{object};
    \end{tikzpicture} \\
     & \cellcolor{bkgray}\begin{tikzpicture}
        \sentence{Someone$_A$ turns Something$_B$ away from Something$_A$};
        \spanannotation{1}{1}{subject};
        \spanannotation{2}{2}{verb};
        \spanannotation{3}{3}{object1};
        \spanannotation{4}{5}{preposition};
        \spanannotation{6}{6}{object2};
    \end{tikzpicture}
    \\ 
    \midrule\multirow{2}{=}[\baselineskip]{2: Emotion or basic human drive that motivates $X$} & \begin{tikzpicture}
        \sentence{Gage wants safety {\scriptsize\relation{Causes/Enables}} Gage turned his bike};
        \spanannotation{1}{1}{subject};
        \spanannotation{2}{2}{verb};
        \spanannotation{3}{3}{object};
        \spanannotation{5}{5}{subject};
        \spanannotation{6}{6}{verb};
        \spanannotation{7}{8}{object};

    \end{tikzpicture}
    \\ 
    & \cellcolor{bkgray}\begin{tikzpicture}[inner sep=0, outer sep=0]
        \sentence{Someone$_A$ wants safety {\scriptsize\relation{Causes/Enables}} Someone$_A$ moves away from Something$_A$ (that is dangerous)};
        \spanannotation{1}{1}{subject};
        \spanannotation{2}{2}{verb};
        \spanannotation{3}{3}{object};
        \spanannotation{5}{5}{subject};
        \spanannotation{6}{6}{verb};
        \spanannotation{7}{8}{preposition};
        \spanannotation{9}{12}{object};
    \end{tikzpicture}
      \\ 
    \midrule\multirow{2}{=}[0.5\baselineskip]{3: Location state that enables $X$} 
    & \begin{tikzpicture}
        \sentence{Gage was close to a car {\scriptsize\relation{Enables}} Gage turned his bike away from the car};
        \spanannotation{1}{1}{subject};
        \spanannotation{2}{2}{verb};
        \spanannotation{3}{4}{preposition};
        \spanannotation{5}{6}{object};
        \spanannotation{8}{8}{subject};
        \spanannotation{9}{9}{verb};
        \spanannotation{10}{11}{object1};
        \spanannotation{12}{13}{preposition};
        \spanannotation{14}{15}{object2};
    \end{tikzpicture}
    \\ 
      & \cellcolor{bkgray}\begin{tikzpicture}
        \sentence{Someone$_A$ ~is~ close to Something$_A$ \relation{\scriptsize Enables} Someone$_A$ moves away from Something$_A$};
        \spanannotation{1}{1}{subject};
        \spanannotation{2}{2}{verb};
        \spanannotation{3}{4}{preposition};
        \spanannotation{5}{5}{object};
        \spanannotation{7}{7}{subject};
        \spanannotation{8}{8}{verb};
        \spanannotation{9}{10}{preposition};
        \spanannotation{11}{11}{object};
    \end{tikzpicture}
      \\ 
    \midrule\multirow{2}{=}[0.5\baselineskip]{4: Possession state that enables $X$}
    & \begin{tikzpicture}
        \sentence{Gage possesses a bike {\scriptsize\relation{Enables}} Gage turned his bike};
        \spanannotation{1}{1}{subject};
        \spanannotation{2}{2}{verb};
        \spanannotation{3}{4}{object};
        \spanannotation{6}{6}{subject};
        \spanannotation{7}{7}{verb};
        \spanannotation{8}{9}{object};
    \end{tikzpicture}
    \\ 
      & \cellcolor{bkgray}\begin{tikzpicture}
    \sentence{Someone$_A$ possesses Something$_A$ {\scriptsize\relation{Enables}} Someone$_A$ moves Something$_A$};
        \spanannotation{1}{1}{subject};
        \spanannotation{2}{2}{verb};
        \spanannotation{3}{3}{object};
        \spanannotation{5}{5}{subject};
        \spanannotation{6}{6}{verb};
        \spanannotation{7}{7}{object};
    \end{tikzpicture}
     \\ 
    \midrule\multicolumn{2}{l}{5: Other attributes enabling $X$: N/A (the dimension is not applicable for this example)}  \\
    
    \midrule
        \multirow{2}{=}[1.1\baselineskip]{6: Event that $X$ directly causes or enables}
    & \begin{tikzpicture}
        \sentence{ Gage turned his bike {\scriptsize\relation{Causes/Enables}} He fell off his bike};
        \spanannotation{1}{1}{subject};
        \spanannotation{2}{2}{verb};
        \spanannotation{3}{4}{object};
        \spanannotation{6}{6}{subject};
        \spanannotation{7}{8}{verb};
        \spanannotation{9}{10}{object};
    \end{tikzpicture}
   \\
     & \cellcolor{bkgray}\begin{tikzpicture}
        \sentence{Someone$_A$ turns Something$_B$ (that is Someone$_A$'s vehicle) {\scriptsize\relation{Causes/Enables}} Someone$_A$ falls off Something$_B$};
        \spanannotation{1}{1}{subject};
        \spanannotation{2}{2}{verb};
        \spanannotation{3}{7}{object};
        \spanannotation{9}{9}{subject};
        \spanannotation{10}{11}{verb};
        \spanannotation{12}{12}{object};
    \end{tikzpicture}
    \\ 
    \midrule\multicolumn{2}{l}{7: An emotion that is caused by $X$: N/A } 
      \\ 
    \midrule\multirow{2}{=}[0.5\baselineskip]{8: A change in location that $X$ results in} 
    & \begin{tikzpicture}
        \sentence{Gage turned his bike away from the car {\scriptsize\relation{Results in}} Gage was further from the car};
        \spanannotation{1}{1}{subject};
        \spanannotation{2}{2}{verb};
        \spanannotation{3}{4}{object1};
        \spanannotation{5}{6}{preposition};
        \spanannotation{7}{8}{object2};

        \spanannotation{10}{10}{subject};
        \spanannotation{11}{11}{verb};
        \spanannotation{12}{12}{object1};
        \spanannotation{13}{13}{preposition};
        \spanannotation{14}{15}{object2};
    \end{tikzpicture}
    \\ 
      & \cellcolor{bkgray}\begin{tikzpicture}
        \sentence{Someone$_A$ moves away from Something$_A$ \relation{\scriptsize Results in} Someone$_A$ ~is~ further from Something$_A$};
        \spanannotation{1}{1}{subject};
        \spanannotation{2}{2}{verb};
        \spanannotation{3}{4}{preposition};
        \spanannotation{5}{5}{object};
        \spanannotation{7}{7}{subject};
        \spanannotation{8}{8}{verb};
        \spanannotation{9}{10}{preposition};
        \spanannotation{11}{11}{object};
    \end{tikzpicture}
      \\ 

    \midrule\multicolumn{2}{l}{9: A change of possession that $X$ results in: N/A} \\
    \midrule\multicolumn{2}{l}{10: Other changes in property that $X$ results in: N/A } \\

    
    \bottomrule
    \end{tabular}
    \caption{Entries in the GLUCOSE dataset that explain the Gage story around the sentence $X$= \textit{Gage turned his bike sharply}. White and gray rows show specific statements and general rules, respectively. The syntactic slots used for constructing each semi-structured entry are shown underneath it.}
    \label{tab:glucose-framework}
\end{table*}

Humans make countless implicit commonsense inferences about everyday situations. For example, consider the following short story from the ROCStories corpus \cite{mostafazadeh-etal-2016-corpus}: \textit{Gage was riding his bike. A car turned in front of him. Gage turned his bike sharply. He fell off of his bike. Gage skinned his knee.} When even young children read this story, they construct a coherent representation of what happened and why, combining information from the text with relevant background knowledge \cite{kintsch1978toward}. For example, they can construct the causal chain that explains how the car's unexpected turn ultimately led to Gage falling, 
describe how Gage's emotion and location changed throughout the story, and even hypothesize that he likely shouted for help after falling. 

Though humans build such mental models with ease \cite{RCSituationModels}, AI systems for tasks such as reading comprehension and dialogue remain far from exhibiting similar commonsense reasoning capabilities. 
Two major bottlenecks have been acquiring commonsense knowledge and successfully incorporating it into state-of-the-art AI systems. To address the first bottleneck, we have built an effective platform to acquire causal commonsense knowledge at scale. To address the second, we show that pre-trained neural models can start to make similar inferences when trained on such rich curated data. 

We introduce the GLUCOSE\footnote{Human brain functions such as thinking, memory, and learning are closely linked to the glucose levels and how efficiently the brain uses this fuel source \cite{glucoseBrain}. If there is not enough glucose in the brain, neurotransmitters are not produced and communication between neurons breaks down. We are calling this resource GLUCOSE, since we believe AI brains need this source of fuel to enable their basic thinking and fill in their reasoning gaps!} (GeneraLized and COntextualized Story Explanations) dataset. Given a short story and a sentence $X$ in the story, GLUCOSE captures ten dimensions of causal explanation related to $X$. These dimensions, inspired by human cognitive psychology, cover often-implicit causes and effects of $X$, including events, location,  possession, and other attributes, the vast majority of which are not captured by existing resources and models. Importantly, GLUCOSE encodes commonsense knowledge in the form of semi-structured inference rules\footnote{We will use ``inference rule'' and ``explanation'' interchangeably: the ``explanations'' we are interested in are inference rules that explain a given sentence's causes and effects.} (mini-theories about the world), each grounded in a specific story. As the examples in Table \ref{tab:glucose-framework} demonstrate, the specific statements exemplify how the general rules can be grounded in a particular context.


To facilitate acquisition at scale, we designed an effective multi-stage crowdsourcing platform and used it to acquire more than 670K GLUCOSE annotations in the context of children's stories. Our analysis shows that these explanations extend substantially beyond the scope of the existing knowledge resources. 


Given the breadth of commonsense knowledge needed for real-world inference tasks, no static knowledge source is expected to provide sufficient coverage. GLUCOSE's key contribution is enabling models to dynamically produce general inference rules to explain novel scenarios. To systematically evaluate such models, we present an evaluation task where given a story $S$, a sentence $X$, and dimension $d$, a model predicts relevant specific and general rules as captured in GLUCOSE. We evaluate on the task using a curated test set, based on novel stories not used for any training purposes. We show a strong correlation between human and automatic evaluation metrics, which makes systematic and reliable evaluation of models feasible. We show that pre-trained neural models perform poorly on the task; however, when finetuned on GLUCOSE data, they are able to generate commonsense explanations that rival humans'. This finding supports our hypothesis that a promising recipe for giving machines commonsense is to use quality-monitored crowdsourced commonsense knowledge for training neural models that have pre-existing lexical and conceptual knowledge.

\section{Related Work}
\label{sec:related-work}

Recently, there has been a renewed interest in commonsense reasoning \cite{talmor-etal-2019-commonsenseqa,tandon-etal-2019-wiqa,rashkin-etal-2018-modeling,zellers-etal-2018-swag}, 
further fostered by the increasing need for explainable AI systems \cite{yang-etal-2018-commonsense}.

One well-known type of commonsense knowledge is script knowledge, defined by \citet{scripts} as structured knowledge about stereotypical event sequences and their participants. However, manual encoding of such knowledge is notoriously unscalable and brittle. A more recent line of work is unsupervised learning of ``narrative schemas'' \cite{chambers-jurafsky-2008-unsupervised,chambers-jurafsky-2009-unsupervised,balasubramanian-etal-2013-generating,sha-etal-2016-joint}, where common event sequences are automatically induced from large corpora. While promising, this approach has not produced high-quality knowledge usable for downstream tasks at scale \cite{mostafazadeh-etal-2016-corpus}. Furthermore, since commonsense knowledge is often implicit, such corpus-based methods are unlikely to induce implicit commonsense inferences \cite{Gordon:2013:RBK:2509558.2509563}. In contrast, our data collection framework has enabled us to acquire high-quality and robust commonsense knowledge, including often unstated rules such as ``Someone$_A$ gives Someone$_B$ Something$_A$  \relation{Results in} Someone$_B$ possesses Something$_A$'' or ``Someone$_A$ is at Somewhere$_A$ \relation{Enables} Someone$_A$ puts Something$_A$ at Somewhere$_A$''. 

The most fruitful efforts to date for acquiring commonsense knowledge have been crowdsourced knowledge resources. ConceptNet \cite{speeretal-aaai17}, a  partially-crowdsourced resource, is a relational knowledge graph that connects short natural-language phrases via semantic edges. Most ConceptNet knowledge is taxonomic, consisting of factoids like ``apple \relation{is a} fruit'', however, it also includes some causal relations, e.g., ``kill \relation{is motivated by} revenge.'' Despite its broad coverage, ConceptNet has been found to be noisy \cite{Zhou:2019:PCP:3308558.3313486}. Its knowledge also lacks context, hampering accurate application at inference time, e.g., ``kill \relation{requires} eat breakfast'' is hard to make sense of without more context.

A more directly relevant resource is ATOMIC \cite{sapetal-aaai19}, which consists of 877K textual descriptions of if-then knowledge. Each entry describes a likely cause/effect of one of 24K+ events. ATOMIC entries are organized into nine categories such as \formal{xIntent} (PersonX's intention) and \formal{xEffect} (effect on PersonX). For instance, ``PersonX makes PersonY’s coffee \formal{xEffect} PersonX gets thanked''. ATOMIC is a great step forward in acquiring high-quality inferential knowledge. However, it has two main shortcomings. 
First, ATOMIC is non-contextual and conflates knowledge about an event that may have occurred under different scenarios, which hinders interpreting and applying the knowledge in context. For example, the event ``PersonX arrives the next day'' has \formal{xIntent}s ``to go on vacation'' and ``to attend a reunion,'' and  \formal{xEffect}s ``get time to relax'' and ``meet some friends.'' Although each \formal{xIntent} should be associated with only one of the \formal{xEffect}s, such dependencies are not encoded in ATOMIC. As a result, ATOMIC cannot be used to determine which \formal{xEffect} is more likely given an \formal{xIntent}. GLUCOSE addresses this by grounding each piece of inferential knowledge to a particular story context consistent across dimensions.

Second, events and relations in ATOMIC are person centric; agentless events are not covered, and each relation is either about PersonX or PersonY. As a result, ATOMIC cannot describe events involving common entity types such as places, things, or groups of people, nor can it encode causes and effects other than to PersonX and their peers. In GLUCOSE, sentence $X$ can describe any event/state, and GLUCOSE general rules can refer to indexed variables such as ``Someone$_A$'' or ``Somewhere$_C$.'' Beyond these major shortcomings, ATOMIC also does not cover many commonsense knowledge types in GLUCOSE, including change of attributes such as location, which will be further discussed in Section \ref{sec:comparisons}. 


\section{The Knowledge Model of GLUCOSE}\label{sec:glucose-framework}
GLUCOSE has a unique take on explaining story events. As illustrated in Table \ref{tab:glucose-framework}, each story is explained through ten causal dimensions. The semi-structured explanation for each dimension includes both a specific statement and a general rule. 

\subsection{Causal Dimensions of Explanation}

One of our main contributions is the identification of ten causal dimensions of explanation in the context of narratives, for which we can reliably collect high quality data from lay crowd workers. Cognitive psychology research on human comprehension of narratives \cite{kintsch1978toward,zwaan1998situation,10.3389/fpsyg.2018.00724} suggests that humans primarily focus on events, their timeline, locations of entities throughout the story, causes and motivations of events, and emotional trajectory of characters. 

Based on this research, GLUCOSE dimensions are designed to focus on causal reasoning around events and states, eliciting event causal chains, character motivations, emotions, naive psychology, and change of attributes such as location and possessions to core story entities. For an event or state $X$ stated in a sentence, we categorize the dimensions of causality into events and states happening \textit{before $X$} and those occurring \textit{after $X$}. Each category includes five dimensions, as shown in Table~\ref{tab:glucose-framework}. The precise definition and scope of these ten dimensions are the result of multiple pilot studies with crowd workers to identify intuitive and distinguishable causal dimensions, so that the overlap among dimensions is minimized and the agreement among workers is maximized.

\subsection{Semi-structured Inference Rules}

To uncover what constitutes a good explanation, we ran several pilot studies exploring how people define, generate, and present explanations about short stories. We concluded that in order to achieve some consensus among explanations and to facilitate further processing and evaluation, the explanations should not be entirely free-form. Instead, we represent them as semi-structured inference rules whose expressivity lies between free text and logical forms. Each rule takes the form ``antecedent \relation{connective} consequent,'' where the antecedent and consequent are composed by filling in syntactic slots for subject, verb, object(s), and preposition(s). For some dimensions, slot-filling involves choosing from a predefined list, e.g., dimension 2, which states a motivating emotion or basic human drive, limits its verb choices to \textit{feel, want}, and \textit{like}. Details regarding the slots can be found in Appendix A.

To eliminate the need for pronoun resolution when applying our general rules, 
variables are indexed, such as ``Someone$_A$'' and ``Something$_A$ and Something$_B$'', to refer to the same entities on both sides of the rule. Each variable can be further elaborated using an \textit{attribute phrase} in the form of a relative clause, e.g., ``Somewhere$_C$ (that is Someone$_A$'s location).'' Our studies indicate that this format gives the explainers sufficient expressivity to convey their reasoning, yet constrains the resulting explanations enough to identify commonalities between them. Note that the semi-structured rules are deterministically converted to natural language form by simply concatenating all the filled slots. Table \ref{tab:glucose-framework} shows examples of semi-structured GLUCOSE explanations.

\subsection{Generalized and Contextualized}
Each GLUCOSE explanation is stated both as a specific statement (grounded in a given context) and a corresponding general rule (applicable to other contexts). Research in cognitive psychology suggests that humans typically choose which of an event's many causes to cite based on its relevance to the context \cite{miller2018insights}. Hence, grounding explanations in context is crucial for acquiring accurate explanations. 
Furthermore, it has been shown that human explanations take situation-specific information and link it to pre-existing knowledge about the world; people explain by appealing to broader theories that enable generalization \cite{Lombrozo2006-LOMTSA}. 
Also, there is evidence that explanations and generalizations help scaffold cognitive development in humans \cite{busch2018explanation}, which can potentially play a role in the learning capabilities of AI systems as well.
By explicitly stating general rules as mini-theories of how the world works, GLUCOSE seeks to enable better generalization and causal reasoning in future AI systems.

\section{The GLUCOSE Dataset} \label{sec:dataset}

\subsection{Data Acquisition Platform}\label{sec:crowdsourcing}

To enable developing models that can build mental models of narratives, we aimed to crowdsource a large, quality-monitored dataset. Beyond the scalability benefits, using crowd workers (as opposed to a small set of expert annotators) ensures diversity of thought, thus broadening coverage of a commonsense knowledge resource.

The annotation task is complex: it requires annotators to understand different causal dimensions in a variety of contexts and to come up with generalized theories beyond the story context. For strict quality control, we designed a three-stage knowledge acquisition pipeline for crowdsourcing the GLUCOSE dataset on the Amazon Mechanical Turk (Mturk) Platform. The workers first go through a qualification test\footnote{GLUCOSE qualification UI: \url{https://bit.ly/34Pej0N}} where they must score at least 90\% on 10 multiple-choice questions on select GLUCOSE dimensions. Next, qualified workers can work on the main GLUCOSE data collection task: given a story $S$ and a story sentence $X$, they are asked to fill in (allowing for non-applicable) all ten GLUCOSE dimensions, getting step-by-step guidance from the GLUCOSE data acquisition UI.\footnote{GLUCOSE main knowledge acquisition UI: \url{https://bit.ly/2R8XcTt}} To ensure data consistency, the same workers answer all dimensions for an $S, X$ pair. Finally, the submissions are reviewed by an expert who rates each worker on a scale from 0 to 3, and provides feedback on how to improve. Our final UIs are the result of more than six rounds of pilot studies, iteratively improving the interaction elements, functionality, dimension definitions, instructions, and examples.\footnote{Our pilot studies helped narrow our dimensions from 18 down to 10 which workers could reliably distinguish. Notably, we collapsed \formal{Enable} and \formal{Cause} on which workers had significant disagreement.} See Appendix B for more details on our crowdsourcing pipeline.\footnote{Additional information about the pipeline and data quality management can be found at \url{https://tinyurl.com/y2pn5cgl}}

\subsection{Dataset Composition and Statistics}\label{sec:stats}
\begin{figure}[tb]
    \begin{center} \includegraphics[width=0.47\textwidth]{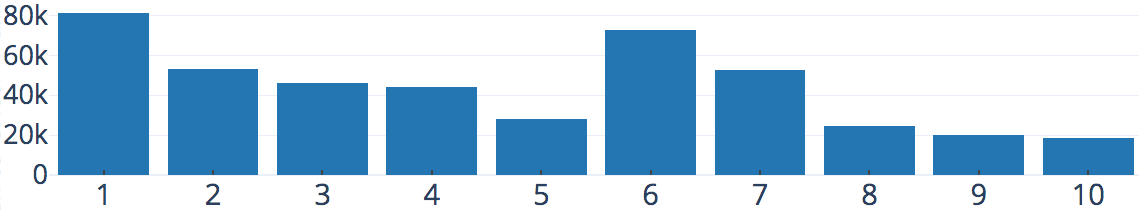}
    \end{center}
    \caption{Number of rules collected for each dimension. Dimensions 1 and 6 have the most representation, while dimensions 9 and 10 are most often marked as not applicable.}
    \label{fig:dimensions-distro}
\end{figure}

Our source of stories for the GLUCOSE dataset is ROCStories \cite{mostafazadeh-etal-2016-corpus}. ROCStories consists of crowdsourced five-sentence everyday stories rich in causal and temporal relations, making them ideal for acquiring commonsense knowledge. We focus on children's stories due to their simpler language and concepts. We computed an estimated target age\footnote{Target age of individual stories was judged by age-of-acquisition and readability tests: Flesch-Kincaid Grade Level, the Coleman-Liau Index, and the Dale-Chall formula \cite{Kuperman2012}. It is important to note that this method depends on vocabulary and does not ensure that all content is appropriate for children in this age group.} for each story and sampled from the 5--8 age group. To ensure diverse viewpoints and hypotheses, each $S, X$ pair was assigned to three workers. Data collection statistics are shown in Table \ref{tab:dataset-stats} and Figure \ref{fig:dimensions-distro}.

\begin{table}[tb]
    \setlength{\tabcolsep}{9.0pt}
    \centering
    \small
    \begin{tabular}{lr}
        \toprule
        \# total annotations  &  \~{}670K\\
        \# total pair of rules  &  \~{}335K\\
        \# total unique stories $S$  &  4,881\\
        \# workers participated &  371\\
        Avg \# of submissions by a worker & 130.7 \\
        Max \# of submissions by a worker & 3,757 \\
        Avg minutes of work time / submission & 8.78 \\
        Avg payment / submission & \$1.60\\
        Avg \# of dimensions filled in / submission & 4.5\\
        \bottomrule
    \end{tabular}
    \caption{Statistics about the GLUCOSE dataset.}
    \label{tab:dataset-stats}
\end{table}


As Figure~\ref{fig:dimensions-distro} shows, the causal dimensions (1 and 6) have the most representation (18.1\% and 16.4\%, respectively). As our examples in Table~\ref{tab:glucose-framework} show, specific statements for these dimensions sometimes define a causal connection over paraphrases of story sentences\footnote{It is important to note that, even if the antecedent and consequent are both in the story, making the causal link between them explicit is considered to have fulfilled the purpose of providing common sense knowledge.}, rather than introduce novel non-story content in either the antecedent or the consequent. To estimate how prevalent this phenomenon is, we manually evaluated 100 random samples of specific rules for each of dimensions 1 and 6. We found that for 66\% and 63\% of the samples, for dimensions 1 and 6 respectively, at least one of the annotators contributed statements that contained inferences with non-story content. The new content includes events that are likely to follow from the story as well as world knowledge about story entities.

\subsection{Comparison to Other Resources}\label{sec:comparisons}

To assess the novelty of GLUCOSE knowledge, 
we compared its coverage  against that of the two most relevant commonsense resources: ConceptNet and ATOMIC.\footnote{Note that \cite{rashkin-etal-2018-modeling} and \cite{rashkin-etal-2018-event2mind} are in essence a subset of ATOMIC, and hence, have even lower coverage compared with GLUCOSE.} We performed a best-effort mapping from GLUCOSE dimensions to relations in ConceptNet and ATOMIC. 
For example, GLUCOSE dimensions 1 and 6 are mapped to ConceptNet's \relation{Causes}, \relation{HasSubevent}, \relation{HasPrerequisite}, and to ATOMIC's \relation{xEffect} and \relation{oEffect}. For all mappings see Appendix A.


Since all three resources contain mostly natural-language entries, it is not possible to automatically quantify their precise overlap, so we adopted a lenient evaluation scheme. For each GLUCOSE general rule\footnote{We evaluated GLUCOSE's specific statements against ConceptNet, with nearly identical results to those in Table~\ref{tab:resource-comparison}.} $A$ \relation{relation} $B$, we queried each target resource for tuples $R'(A', B')$, where $R'$ is the resource's mapped equivalent of \relation{relation}, and $A'$ and $B'$ consist of just the main verbs in $A$ and $B$. Using fuzzy matching on $A'$ and $B'$, we retrieved a large number of hits for the query, then filtered to those with $>$50\% lexical overlap with the GLUCOSE rule.

The results, shown in Table~\ref{tab:resource-comparison}, represent a ceiling in overlap with other resources. The results indicate that GLUCOSE captures extensive commonsense knowledge unavailable in existing resources. Note that GLUCOSE’s knowledge model is a superset of ATOMIC's. GLUCOSE is designed to encompass all nine categories of inferential commonsense knowledge that ATOMIC covers, which are captured across different GLUCOSE dimensions. Note that there are definitely some individual pieces of knowledge that have been acquired in ATOMIC which do not exist in GLUCOSE, since some ATOMIC events may not have appeared in the GLUCOSE stories.


\begin{table}[tb]
    \centering
    \small
    \setlength{\tabcolsep}{4.4pt} 
    \begin{tabular}{lcccccc}
        \toprule
        Dimension & 1 & 2 & 5 & 6 & 7 & 10 \\\midrule
        ConceptNet &  1.2\% & 0.3\% & \phantom{0.}0\% & 1.9\% & \phantom{0.}0\% & \phantom{0.}0\%\\
        ATOMIC    & 7.8\% & 1.2\% & 2.9\% & 5.3\% & 1.8\% & 4.9\%\\
        \bottomrule
    \end{tabular}
    \caption{Ceiling overlap between GLUCOSE and other resources. Omitted dimensions had no overlap.}
    \label{tab:resource-comparison}
\end{table}

\section{Empirical Evaluation Task}\label{sec:task}

We set up a standalone evaluation task for evaluating models that predict GLUCOSE explanations: given a story $S$, a story sentence $X$, and a dimension $d$, provide an explanation in both specific and general forms.

\paragraph{Test Set Curation}  For a test set on commonsense reasoning to offer accurate and reliable evaluation, it should contain unambiguous examples with clear gold answers. This led to a curation process that identifies examples on which humans have high agreement, as follows: we sampled $S, X$ pairs annotated by any three workers with the highest quality rating. A dimension $d$ for $S, X$ was allowed into the test set if 1)
$d$ was annotated by all three workers, and 2) the three specific statements had a round-robin average sentence-level BLEU \cite{lin-och-2004-automatic} score\footnote{
We averaged the BLEU scores obtained, in round-robin fashion, by taking one rule as candidate and the other two as references. We used BLEU with equal weights up to 4-grams.} above 0.75. Finally, two in-house annotators manually removed cases with typographical or core content errors, resulting in a test set of 500 story/sentence pairs, each with 1-5 dimensions answered.
\paragraph{Human and Automatic Evaluation} Human evaluation is crucial for any language generation task. We crowdsourced our human evaluation on MTurk, using a dedicated UI,\footnote{GLUCOSE evaluation UI: \url{https://bit.ly/2rJWFwy}} asking three of our top-rated crowd workers from the main GLUCOSE crowdsourcing job to rate the predictions. We set up the following evaluation process to ensure calibrated judgments: the judge first reads a story with a highlighted sentence $X$, then reads a question about $X$ corresponding to a GLUCOSE dimension. Next, they are shown a shuffled list of candidate answers, each produced by a different system. Finally, the judge rates each candidate answer on a four-point Likert scale: ``completely incorrect,'' ``almost incorrect,'' ``almost correct,'' and ``completely correct.'' To compare system performance, the ratings are mapped to numerical scores of 0--3, which are then averaged. 


Automatic evaluation for tasks involving language generation has been a major bottleneck for research \cite{liu-etal-2016-evaluate,hashimoto-etal-2019-unifying}. BLEU's ease of replicability has made it a popular automated metric, but its correlation with human judgement has proven weak on various tasks \cite{novikova-etal-2017-need,Gatt:2018:SSA:3241691.3241693}.  For automatic evaluation, we use SacreBLEU \cite{post-2018-call} with equal weights up to 4-grams at corpus-level on the three-reference test set. Using pairwise correlation analysis, we found strong correlation between human and BLEU scores on our test set, with correlation coefficients Spearman = 0.891, Pearson = 0.855, and Kendall's $\tau$ = 0.705, all with $p$-value $<$ 0.001. 
The high correlation is due to various design choices, including 1) semi-structured inference rules in GLUCOSE are designed to be evaluable, where the structure constrains the variability of the rules, and 2) we minimized the noise in our human evaluation by designing a UI that could collect calibrated ratings from human judges educated about the task. The strong correlation suggests that BLEU is a viable metric for reporting future results on the GLUCOSE test set. 


\section{Models}\label{sec:models}

We developed several models for tackling the prediction task described in Section \ref{sec:task}. The train and development sets for each model consisted of the initial 440K total annotations\footnote{Table \ref{tab:dataset-stats} shows the statistics of the final dataset, whereas all training for the models in the paper were conducted before the crowdsourcing of the dataset was finished.} (in the context of 3,360 stories) in the GLUCOSE dataset, minus the entries that share the context story with the test instances.


Due to their superior performance in sequence prediction, all our neural models use transformer blocks \cite{NIPS2017_7181}, which use multi-headed attention and fully connected layers to encode sequences. For decoding, all models use top-k random sampling \cite{fan-etal-2018-hierarchical}. Details on all the models we experimented with can be found in Appendix C.

\subsection{Pretrained Language Model (PT-LM)} PT-LM tests what GLUCOSE-like knowledge is captured by the pretrained 774M-parameter GPT-2 \cite{radford2019language} language model. We elicit commonsense explanations from GPT-2 by prompting it with the story followed by sentence $X$ and a dimension-specific trigger word like ``because'', and allowing the model to complete the sentence. For best results, we implemented ``constrained decoding'' by conditioning the GPT-2 model on the input $S,X$ as context, then generating the next token for a dimension $d$ as follows: if dimension $d$'s template specifies a set of allowable words at the current position---e.g., locative prepositions for dimensions 3 and 8---sample from the options based on their likelihood as conditioned on the preceding tokens. Otherwise, allow sampling freely from the entire vocabulary. See Appendix C for a list of all templates used.

\subsection{Models Trained on GLUCOSE}
 
 

\subsubsection{Language Models}
We finetuned separate language models for specific and general rules. Each model monolithically covers all ten GLUCOSE dimensions: it generates rules given a dimension indicator as input.\footnote{We experimented with training separate models for each dimension, which yielded much worse results.} Rules are sampled from the learned distribution $p(s)=\prod_{i=1}^{n}p(s_i \mid s_1, \ldots, s_{i-1})$, where $s$ is the concatenation of input and output sequences. For all models in this section, we finetuned the PT-LM model described above. 


\paragraph{One-sided Generation (1S-LM)} One side of a GLUCOSE rule---the antecedent or the consequent, depending on the dimension---is always a paraphrase and/or a generalization of sentence $X$. In the one-sided model, we use $X$ as is for this side of the specific statement; the model generates only the \textit{target} side. Each training example is a text sequence \textit{$S$\#$X$\#$d$\#answer\#EOS}, where $d$ is the dimension number and \textit{answer} is the target side. At test time, the model generates answer characters until it produces an EOS token. 



\paragraph{Full Rule Generation (Full-LM)} 
Full-LM learns to produce the complete rule, including the connective and the paraphrase of $X$. Instead of just the target side of the rule, the training examples have the full rule as the \textit{answer} portion of the sequence. This allows the model to produce more human-like rules, including paraphrasing and/or generalizing $X$ appropriately.  


\begin{table*}[tb]
    \small
    \centering
    \setlength{\tabcolsep}{3pt}    \newcommand{\humanbleusep}{15pt}
    \begin{tabular}{=l +c +c +c +c +c +c +c +c +c +c @{\hspace{\humanbleusep}} +c +c +c +c +c +c +c +c +c +c}
        \toprule
        & \multicolumn{10}{c}{\textbf{Human evaluation scores for dimension...}} & \multicolumn{10}{c}{\textbf{BLEU scores for dimension...}} \\ \cmidrule(lr{15pt plus \cmidrulekern}){2-11} \cmidrule(lr){12-21}
        \rowstyle{\bfseries} Model & 1 & 2 & 3 & 4 & 5 & 6 & 7 & 8 & 9 & 10 & 1 &  2 & 3 & 4 & 5 & 6 & 7 & 8 & 9 & 10 \\
        \midrule
        PT-LM & 0.7 & 1.0 & 1.2 & 1.0 & 0.6 & 0.6 & 0.6 & 0.9 & 0.7 & 1.1 & 40.7 & 36.5 & 31.3 & 31.4 & 30.2 & 32.1 & 23.1 & 37.0 & 40.9 & 53.1 \\ \midrule
        1S-LM & 2.1 & 2.3 & 2.2 & 2.5 & 2.1 & 2.1 & 2.4 & 2.5 & 2.1 & 1.8 & 55.1 & 59.6 & 50.7 & 65.2 & 53.1 & 57.4 & 55.4 & 71.7 & 56.8 & 67.2 \\
        \midrule
        \multirow{2}{*}{Full-LM} & 1.8 & 2.0 & 2.0 & 2.2 & 1.7 & 2.0 & 2.1 & 2.2 & 1.6 & 2.1 & 54.7 & 55.3 & 51.0 & 64.4 & 50.5 & 58.8 & 66.2 & 73.4 & 32.7 & 67.0 \\
        & \rowstyle{\cellcolor{bkgray}} 1.6 & 1.6 & 1.8 & 2.1 & 1.8 & 1.9 & 1.9 & 2.1 & 1.1 & 1.5 & 56.4 & 55.8 & 57.5 & 62.7 & 59.6 & 59.0 & 65.8 & 67.7 & 53.7 & 56.2 \\ \midrule
        \multirow{2}{*}{Enc-Dec} & \rowstyle{\bfseries} 2.7 & 2.7 & 2.6 & 2.7 & 2.5* & 2.6 & 2.7 & 2.8 & 2.2 & 2.5* & 72.5 & 73.9 & 73.8 & 79.3 & 70.5 & 80.2 & 81.1 & 86.6 & 71.7 & 66.9 \\
        & \rowstyle{\cellcolor{bkgray}\bfseries} 2.3 & 2.3 & 2.4 & 2.5 & 2.3 & 2.4 & 2.5 & 2.7 & 1.9 & 1.7* & 66.4 & 67.6 & 68.5 & 73.0 & 69.8 & 77.6 & 76.8 & 86.8 & 68.6 & 57.5 \\ \bottomrule
        \multirow{2}{*}{Human} & 2.8 & 2.7* & 2.8 & 2.9 & 2.5* & 2.8 & 2.8 & 2.8 & 2.9* & 3.0 &&&&&N/A&&&&& \\
        & \rowstyle{\cellcolor{bkgray}} 2.5 & 2.6 & 2.4 & 2.6 & 2.4 & 2.6 & 2.6 & 2.6 & 2.6* & 2.7 &&&&&N/A&&&&& \\ \bottomrule
    \end{tabular}
    \caption{Evaluation results for GLUCOSE models. Human evaluation scores are out of 3; BLEU scores are out of 100. Gray and regular rows show results on general and specific rules, respectively. Human model's performance was computed by showing judges a randomly selected answer from the three gold references.
    We performed paired sample t-tests on the human evaluation scores for each dimension for Full-LM against Enc-Dec, and then again for Enc-Dec against Human. The vast majority of differences are statistically significant at $p<0.05$, with the exceptions noted in asterisk. Note that the dimensions where performance differences are not statistically significant strongly correlate with those with the least amount of data, as shown in Figure~\ref{fig:dimensions-distro}.
    }\label{tab:results}
\end{table*}

\begin{table*}[tb]
    \centering
    \small
    \begin{tabular}{=p{2.8em}+p{0.34\linewidth}+p{0.51\linewidth}}
        \toprule
         \bf Model & \bf Dim 3: A location state that \relation{Enables} $X$ & \bf Dim 6: An event that $X$ \relation{Causes/Enables}  \\ \midrule
        \multirow{4}{=}{Full\hyp{}LM} &Karen  is at home \relation{Enables}  Karen made  a pan of lasagna  and brought it  to  the party &Karen made lasagna \relation{Causes/Enables} Karen ate lasagna
        \\
        & \rowstyle{\cellcolor{bkgray}} Someone$_A$ is in Somewhere$_A$ \relation{Enables} Someone$_A$ makes Something$_A$ (that is edible)  & Someone$_A$ cooks Something$_A$ (that is food) \relation{Causes/Enables} Some People$_A$ to be turned away because of  Something$_A$ (that is food)\\
        \midrule
        \multirow{4}{=}{Enc\hyp{}Dec} & Karen is in the kitchen \relation{Enables} Karen makes a pan of lasagna &Karen makes a pan of lasagna \relation{Causes/Enables} Karen eats it for a week \\
         & \rowstyle{\cellcolor{bkgray}} Someone$_A$ is in a kitchen \relation{Enables} Someone$_A$ cooks Something$_A$ & Someone$_A$ makes Something$_A$ (that is food) \relation{Causes/Enables}  Someone$_A$ eats Something$_A$ \\
        \midrule
        \multirow{4}{=}{Human} & Karen is in the kitchen \relation{Enables} Karen made a pan of lasagna & Karen made a pan of lasagna \relation{Causes/Enables} She brought it to a party\\
        & \rowstyle{\cellcolor{bkgray}} Someone$_A$ is in a kitchen \relation{Enables} Some\-one$_A$ prepares Something$_A$ (that is a dish) & Someone$_A$ prepares Something$_A$ (that is a dish) \relation{Causes/Enables} Someone$_A$ takes Something$_A$ to  Something$_B$ (that is an event)\\
        \bottomrule
    \end{tabular}
    \caption{Example model generations for the input story: \textit{{\underline{Karen made a pan of lasagna.} She brought it to the party. Nobody wanted to eat lasagna. Karen ate it for a week. She became tired of lasagna.}} (Sentence $X$ is underlined.) Note that all test stories are unseen in the train or validation set.}\label{tab:examplegen}
\end{table*}


\subsubsection{Encoder-Decoder Model (Enc-Dec)}
Our most complex model is an encoder-decoder transformer model that jointly predicts the specific and general rules. It maximizes $p(y \mid x) = \prod_{i=1}^{n}p(y_i \mid x; y_1,\ldots,y_{i-1})$, where $x$ is the input and $y$ is the answer. We obtained the best results by formulating the input as \textit{\#$d$: $S^*[X]$}, where $d$ is the dimension and $S^*[X]$ is the story $S$ with sentence $X$ surrounded by asterisks. We chose to finetune the state-of-the-art T5 model (with 770M-parameters, to be comparable to the size of the LM model), using the same hyperparameters as in \citep{2019t5}.

\section{Results and Discussion}\label{sec:results}
Table \ref{tab:results} shows the results from the models described in Section~\ref{sec:models}, evaluated as per Section~\ref{sec:task}. It shows that Enc-Dec uniformly outperforms all other models, confirming that full visibility into context\footnote{A clear drawback of language models is that the model's representation of the $i$th item depends only on items preceding $i$, and not the full input context. We show that better predictions can be made given full visibility into the entire input sequence.} helps an architecture better learn the intricacies of GLUCOSE rules.

In fact, Enc-Dec performs competitively with humans in many dimensions. The strength of this model's performance in predicting both specific and general rules is a testament to the high quality of the GLUCOSE training data. Its worst performance is on general rules for dimensions 5 and 10, which have the lowest number of training points and are the most diverse in content. 

Other models perform as expected. PT-LM's poor performance shows that finetuning on our dataset significantly improves the commonsense inference capabilities of LMs. 1S-LM, which only predicts half of an inference rule, outperforms Full-LM in predicting specific statements, but lacks the ability to generalize them. We also tested various other baselines, including an ATOMIC-trained transformer model \cite{bosselut-etal-2019-comet}, retrieval of K-nearest-neighbors, and non-contextual variants of the presented models, all of which significantly underperformed the results in Table \ref{tab:results}, and are presented in Appendix C.

Our results also show that our best models perform noticeably better on specific statements than on general rules. This is because generating a specific statement involves paraphrasing a story sentence and predicting an antecedent/consequent, while a general rule requires further generalizing the paraphrase and the antecedent/consequent appropriately such that the rule remains a generally valid statement about the world.

Although rule generalization can sometimes be as simple as replacing a named entity (e.g., {\em Gage}) with a typed variable ({\em Someone$_A$}), more often more complex transformations are needed, such as generalizing the action and producing type constraints on variables in the form of attribute phrases. For example, take into account the Enc-Dec results in Table \ref{tab:examplegen}. For dimension 3, the generalization of the story sentence, {\em Karen makes a pan of lasagna}, included generalizing {\em Karen} to {\em Someone$_A$} and {\em makes a pan of lasagna}
to {\em cooks Something{$_A$}}. Note that sentence generalizations are dimension-specific: For dimension 6, the generalization of same sentence retains the verb {\em make} but adds a type constraint to the object, {\em Something$_A$ (that is a food)}, which is required for making the rule generally valid. Table \ref{tab:glucose-framework} shows another complex transformation example where {\em turning his bike} is generalized into {\em moves away from Something (that is dangerous)}, that takes into account story context.

 Overall, our evaluation results show that the state-of-the-art pre-trained models finetuned on the GLUCOSE dataset are well capable of dynamically producing GLUCOSE-like inference rules on the fly, which is the ultimate usecase of the GLUCOSE dataset. It is important to note that there is still a consistent performance gap between the best-performing model and human's on generating specific statements and general rules, which indicates that there is still a large headroom for improvement on designing better models for generalizable commonsense reasoning. 
 
 Note that in our current evaluation setup, we have made the simplifying assumption of evaluating each dimension for each sentence individually, without consideration for consistency across dimensions or across sentences. Joint prediction of all the dimensions and sentences across the story is a considerably more challenging task that can potentially yield more accurate predictions for a downstream task. We encourage the future work to focus on building models that perform joint predictions, which can be readily evaluated using our test-set. It is important to note that static test sets are inherently narrow and prone to hidden curation biases \cite{sharma-etal-2018-tackling,belinkov-etal-2019-adversarial}. We believe that the ultimate evaluation for models that show GLUCOSE-like commonsense reasoning capabilities should be on naturally-occurring arbitrary stories and through our presented human evaluation process.  As future work, we are planning to show the value of incorporating GLUCOSE-trained models in other downstream NLP tasks such as reading comprehension and dialog. 

\section{Conclusions}
We introduced GLUCOSE, a large-scale dataset of implicit commonsense knowledge, encoded as explanatory mini-theories grounded in a narrative context. The theories are categorized into ten causal dimensions, inspired by cognitive psychology. 

We presented our multi-stage pipeline for acquiring semi-structured causal explanations at scale from lay workers, 
resulting in \~{}670K annotations in the context of everyday children's stories. 
We demonstrated the utility of GLUCOSE data in two ways. First, our analysis showed that GLUCOSE rules capture knowledge not available in existing resources or pre-trained models. Second, in order to evaluate how well AI models can predict GLUCOSE knowledge on novel inputs, the ultimate value of such a dataset, we defined a standalone evaluation task for predicting specific and general inference rules given a story/sentence pair and a dimension. We curated a doubly-vetted test set, developed a platform to facilitate human judgment of system outputs, and validated BLEU as a strong automated evaluation metric. We show that training on GLUCOSE data improves model performances significantly on unseen stories.

Our results validate our hypothesis that a promising approach for imbuing machines with commonsense is to use carefully-crafted data, as in GLUCOSE, to train neural architectures that have a wide range of lexical and conceptual knowledge encoded, as in models pretrained on large corpora. Together with this paper, we release our dataset\footnote{The GLUCOSE dataset is available for download at \url{https://tinyurl.com/yyeo92pt}.} and models\footnote{The trained models and the details on the GLUCOSE data files can be found through \url{https://github.com/ElementalCognition/glucose/}.}, which we hope will enable the AI research community to explore effective approaches to incorporate commonsense reasoning capabilities into various downstream tasks.

\section*{Acknowledgments}
We would like to thank the hundreds of amazing crowdworkers whose dedication made this work possible. We thank David Ferrucci for his valuable insights and support throughout the GLUCOSE project. We thank Andy Beck for the discussions around the GLUCOSE knowledge model and Jesse Dunietz for his discussions on the paper. We are grateful for the invaluable comments of the anonymous reviewers, Niranjan Balasubramanian, and Owen Rambow on this paper.

\bibliography{anthology,nonacl}
\bibliographystyle{acl_natbib}

\clearpage

\section*{Appendix A: The Knowledge Model for Collecting GLUCOSE data}

\subsection*{Semi-structured Inference Rules }
The knowledge represented in GLUCOSE is captured in the form of semi-structured inference rules that are accompanied by a specific statement that grounds the rule in the context of a specific story. Each specific statement and its corresponding general rule use the common template of \textit{antecedent} \textit{connective} \textit{consequent}. The antecedent and consequent are each composed by filling in a few syntactic slots, namely, subject, verb, object(s), and preposition(s). In order to further shape the semantics of the acquired knowledge, some of these slots have a pre-defined list of options to choose from. 

Table \ref{tab:slots} lists the pre-defined options for filling in the syntactic slots per GLUCOSE dimension\footnote{A sample of the semi-structured rules in GLUCOSE can be found through \url{https://bit.ly/2LFuwOt}.}. Some of the slots allow adding a custom entry to the list of options, hence soft constraints, and some do not, hence hard constraints. Note that beyond the options listed in this table, the general rule slots across all the dimensions have pre-defined options for subject and object slots such as \textit{Someone$_A$} or \textit{Some People$_C$}.

\subsection*{Comparison to Other Resources}
To assess the value of the GLUCOSE dataset, we compared its coverage against the two most relevant commonsense knowledge resources: ConceptNet and ATOMIC. Table \ref{tab:comparison} shows our best-effort mapping among knowledge dimensions of GLUCOSE and relations in ConceptNet and ATOMIC.



\section*{Appendix B: Data Collection Pipeline}

To ensure obtaining our desired quality, we designed a three-stage knowledge acquisition pipeline for crowdsourcing the GLUCOSE dataset on the Amazon Mechanical Turk (Mturk): The qualification test, the main task, and the expert review. In this Section we provide more detail about each stage and its designated UI design. 

\begin{figure*}
\centering
    \includegraphics[width=12cm]{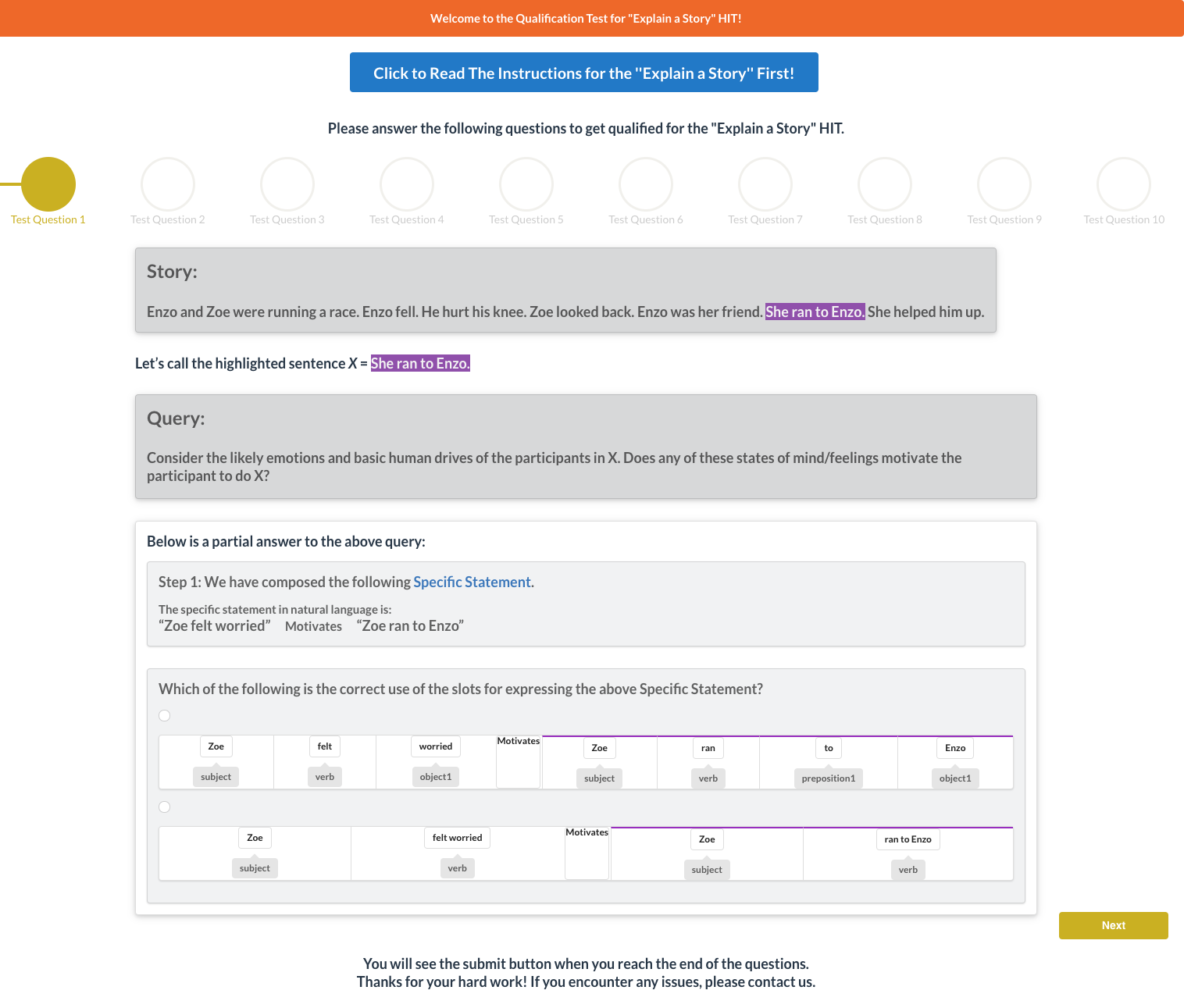}
    \caption{Example qualification question about the correct use of the slots.}
    \label{fig:qual-ui-slots}
\end{figure*}

\begin{figure*}
\centering
    \includegraphics[width=13.5cm]{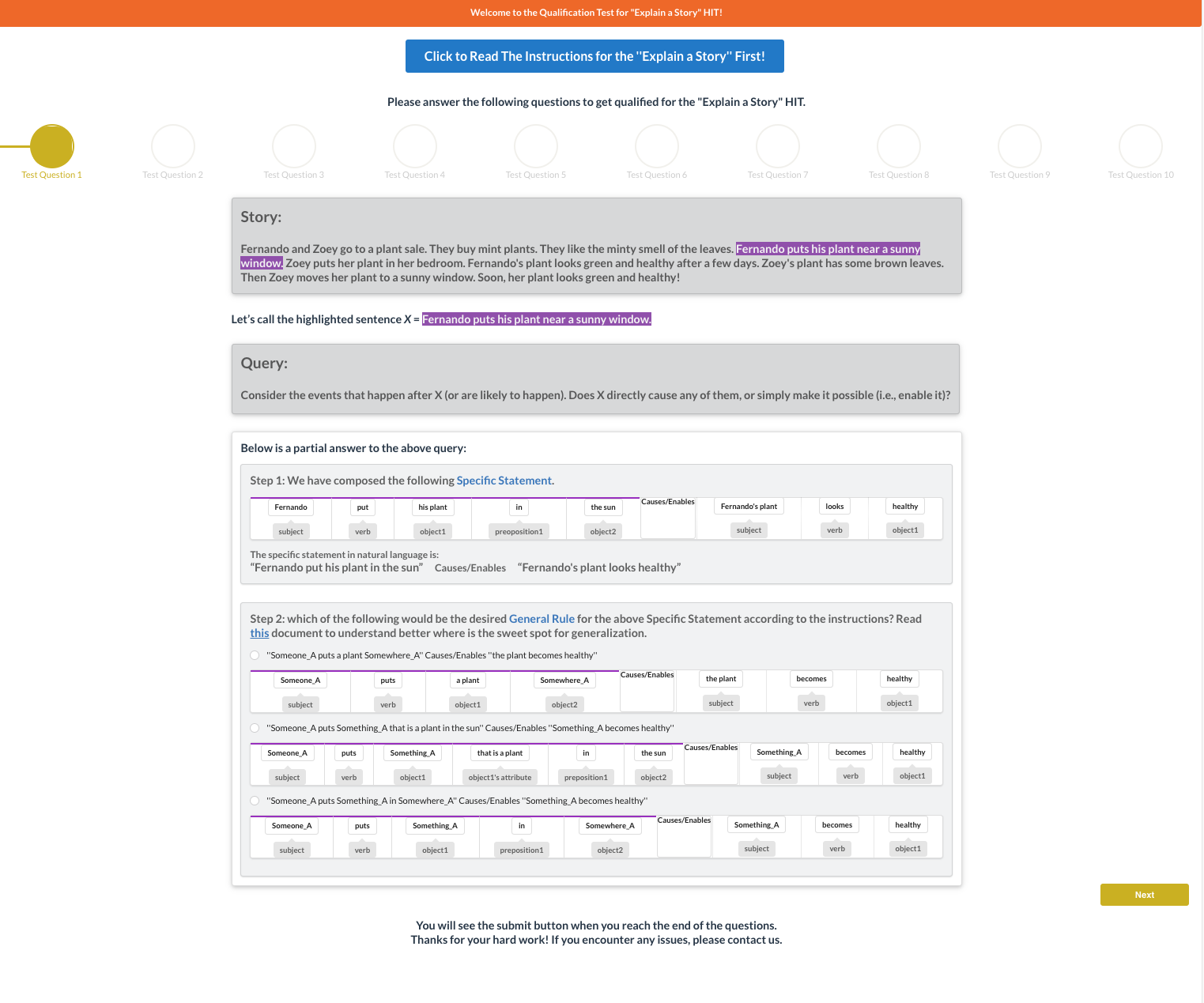}
    \caption{Example qualification question about the correct level of generalization.}
    \label{fig:qual-ui-generalization}
\end{figure*}

\begin{figure*}
\centering
    \includegraphics[width=13cm]{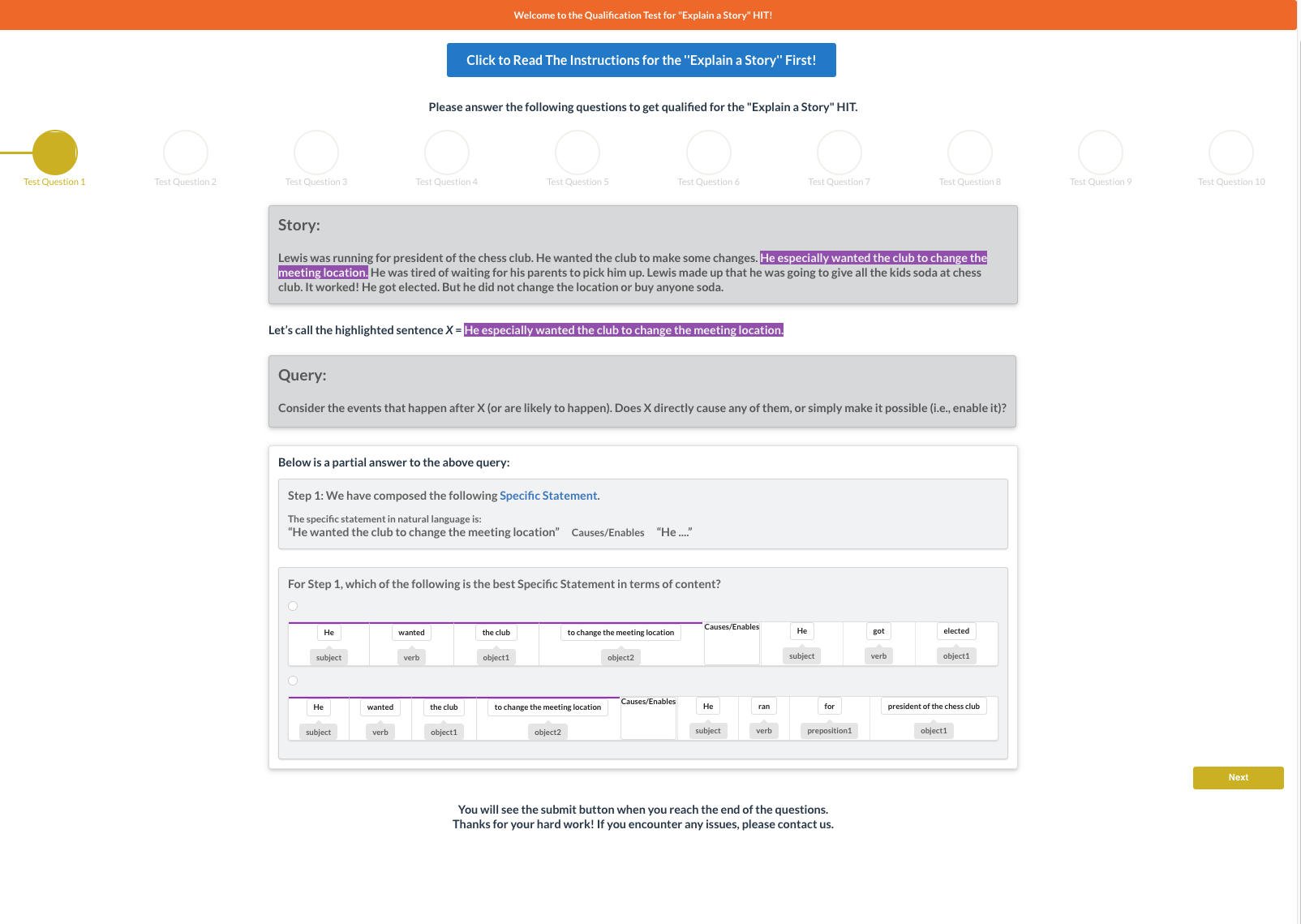}
    \caption{Example qualification question about understanding causal relations between events.}
    \label{fig:qual-ui-conceptual}
\end{figure*}

\paragraph{Qualification Test} The qualification test contained questions testing workers' understanding in three areas: Identifying correct use of the UI slots for composing their answers (Figure \ref{fig:qual-ui-slots}), recognizing the right level of generalization (Figure \ref{fig:qual-ui-generalization}), and identifying causes and effects with proper temporal understanding of the stories (Figure \ref{fig:qual-ui-conceptual}). Understanding generalization is the most difficult, and the most important, aspect of our task. Assessing the  prospective workers' understanding of generalization was done through curating questions demonstrating under-generalization or over-generalization. The full Qualification UI, along with all the detailed instructions that were visible to the workers, is accessible here \url{https://bit.ly/34Pej0N}.





\begin{figure*}
\centering
    \includegraphics[width=10cm]{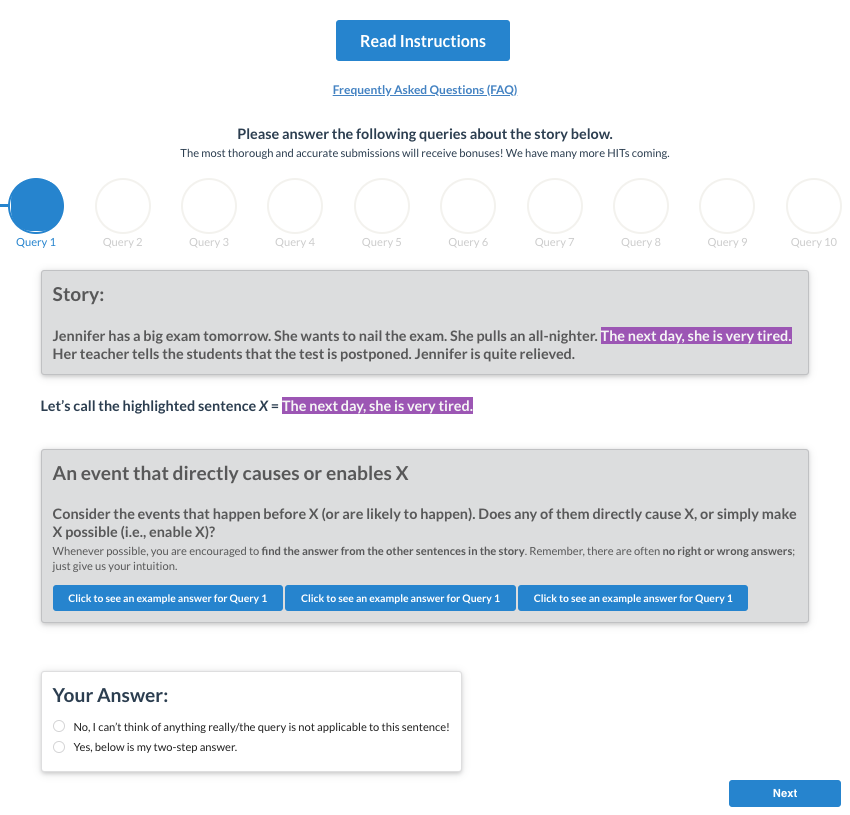}
    \caption{The preview page of the Main UI for GLUCOSE data collection, which can be accessed via \url{https://bit.ly/2R8XcTt}.}
    \label{fig:main-ui}
\end{figure*}

\begin{figure*}
\centering
    \includegraphics[width=12cm]{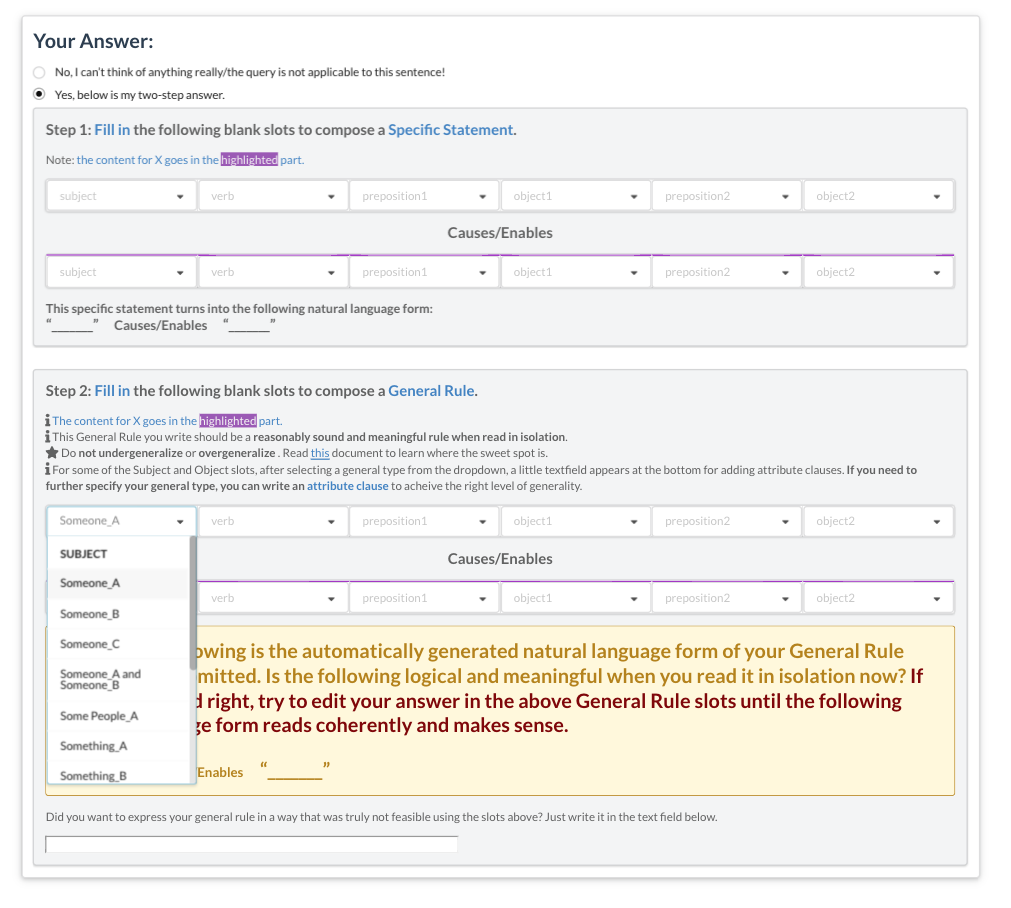}
    \caption{The answer-entry part of the main UI. When ``Yes'' is selected for ``Your Answer'' on the main UI for GLUCOSE data collection, the workers can input answers to the dimension in question.}
    \label{fig:main-ui-slots}
\end{figure*}

\paragraph{Main Task} The qualified workers were able to access large batches of data with no limit. The main task starts with a page like the one shown in the Figure \ref{fig:main-ui}. The user loops through each of the 10 dimensions of GLUCOSE data collection, in order, presented as questions. Note that the user could answer the question by simply marking the dimension as not applicable and skipping it. If they choose to answer, as shown in Figure \ref{fig:main-ui-slots}, they will be presented with the structured rule slots to input their answers. The full Main GLUCOSE UI, along with all the detailed instructions that were visible to the workers, is accessible here \url{https://bit.ly/2R8XcTt}.


%


\begin{table*}[htb]
\begin{center}
\footnotesize
\begin{tabular}{p{5.7cm} p{1.8cm} p{7.5cm}}
\toprule
\bf Dimension & \bf Connective & \bf Slot Constraints  \\
\hline
Dim 1: An event that directly causes or enables X & Causes/Enables & None\\
  & &\\
\hline
Dim 2: An emotion or basic human drive that motivates X & Motivates &\textit{Verb slot hard constraints}: feels, wants, likes;
\textit{Object slot soft constraints}: curiosity, independence, competition, honor, approval, power, status, romance, success, 
friendship, belonging, health, safety, livelihood, 
happy, stressed, angered, disgusted, sad, surprised, 
fearful, trusting, love, obedient, 
amazed, disappointment, regret, worthless, 
aggression, optimistic.
\\
\hline
Dim 3: A location state that enables X & Enables & \textit{Verb slot hard constraints}: am, is, are; \textit{Preposition slot hard constraints}: above, across from, at, below, far from, in, in front of, inside of,near, next to, on top of, outside of. \\
\hline
Dim 4: A possession state that enables X & Enables &  \textit{Verb slot hard constraints}: possess(es).\\
 & &\\
\hline
Dim 5: Other attribute that enables X& Enables & \textit{Verb slot hard constraints}: am, is, are, has, have, want, wants, need, needs. \\
\hline
Dim 6: An event that is directly caused or enabled by X & Causes/Enables & None\\
\hline
Dim 7: An emotion that is caused by X  & Causes & \textit{Verb slot hard constraints}: feels, wants, likes; \textit{Object slot soft constraints}: curiosity, independence, competition, honor, approval, power, status, romance, success, friendship, belonging, health, safety, livelihood, happy, stressed, angered, disgusted, sad, surprised, fearful, trusting, love, obedient, amazed, disappointment, regret, worthless, aggression, optimistic.\\
\hline
Dim 8: A change of location that X results in & Results in &   \textit{Verb slot hard constraints}: am, is, are;  \textit{Preposition slot hard constraints}: above, across from, at, below, far from, in, in front of, inside of,near, next to, on top of, outside of. \\
\hline
Dim 9: A change of possession that X results in & Results in &  \textit{Verb slot hard constraints}: possess(es) \\
\hline
Dim 10: Other change in attribute that X results in & Results in &  \textit{Verb slot hard constraints}: am, is, are, has, have, want, wants, need, needs. \\
\bottomrule
\end{tabular}
\end{center}
\caption{The list of pre-defined options for filling in the syntactic slots per GLUCOSE dimension.}\label{tab:slots}
\end{table*}

\begin{table}[!htb]
\centering
\footnotesize
\begin{tabular}{l l l}
\toprule
\bf Glucose & \bf ConceptNet Rel & \bf ATOMIC Rel \\ \hline
Dims 1 & HasSubevent & xEffect/oEffect\\
\& 6 & HasFirstSubevent & \\
& HasLastSubevent & \\
& HasPrerequisite & \\  \hline
Dim 2  & Desires &  xAttr (``feels")\\
& CausesDesire & xIntent (otherwise)\\ 
& MotivatedByGoal & \\ \hline
Dim 7  & Same as dim2 &  xReact/oReact (``feels") \\ \hline
Dims 5 & Desires & xAttr/xWant \\ 
\& 10 & CausesDesire & oWant \\
\bottomrule
\end{tabular}
\caption{Mappings between GLUCOSE dimensions and ConceptNet/ATOMIC relations. ConceptNet ``Causes'' applies to all GLUCOSE dimensions. Omitted GLUCOSE dimensions have no mapping in ATOMIC.}
\label{tab:comparison}
\end{table}

\paragraph{Expert Review} For work contributed through the main UI, data quality was controlled through daily monitoring of a percentage of incoming submissions and statistics on average dimensions filled out. For managing this process, we built a specialized UI for reviewing the incoming structured data. The percentage of answers reviewed by an in-house expert were used to update worker ratings. Workers enter the task with a score of ``1'', then advance to ``2'' as they become more proficient, getting a bonus increase. The top rating is ``3''. Select workers with a ``3'' rating were also moved into ``top rated'' batches that paid more per HIT and included higher bonuses and incentives. If work quality dropped, workers' ratings were adjusted accordingly. If their work was at a risk of degrading the quality of the dataset, they were disqualified from the task.\footnote{Additional information on the data and data quality management can be found at \url{https://tinyurl.com/y2pn5cgl}.}
\section*{Appendix C: Details on the Models}

\subsection*{ATOMIC-trained Model}
This model is a transformer language model, specifically GPT-1 architecture, fine-tuned on ATOMIC resource. The language model is fine-tuned to generate triplet sequences such as `PersonX goes to the mall \textless xIntent \textgreater to buy clothes'. We use the same exact model trained for \cite{bosselut-etal-2019-comet}. This model is only applicable to General Rule prediction. The results from this model were significantly worse than the PT-LM model, which is the worst-performing model presented in the main paper. This was expected, given the little overlap that exists between the ATOMIC dataset and the GLUCOSE knowledge, as presented in the main paper under "Comparison to Other Resources" Section. 
\subsection*{K-Nearest Neighbor (KNN)} For a given test pair $S, X$, the KNN baseline retrieves the $K$ most similar training instances and returns one as the prediction.  It uses BERT \cite{devlin-etal-2019-bert} sentence embeddings to compute cosine similarity between a candidate and each retrieved training instance. We tuned three parameters on the development set: $K$, $min\_sim$, and $max\_sim$. If a candidate has a similarity score above $max\_sim$, it is emitted as the prediction. Otherwise, candidates scoring below $min\_sim$ are dropped, and the centroid among the remaining pool is emitted. We evaluate KNN only for general rules, since it is not meaningful to retrieve specific statements from the training set. The results from this model were significantly worse than the PT-LM model, which is the worst-performing model presented in the main paper. The performance of the KNN model highlights the importance of generalizing beyond the training data. 

\subsection*{Pretrained Language Model (PT-LM)}
We experimented with prompting the pretrained language models, specifically GPT-2, as is, for predicting GLUCOSE dimensions. Table \ref{tab:decoding} shows the list of particular templates used for decoding. We used 774M-parameter GPT-2 model, with top-K random sampling for decoding, with K = 15. The decoding for this model was done on CPU.

\begin{table*}
\begin{center}
\footnotesize
\begin{tabular}{p{5cm}cl}
\toprule
\bf Dimension & \bf Connective & \bf Natural Language Template  \\
\hline
Dim 1 & Causes/Enables & [because, since]\\
 An event that directly causes or enables X & &\\
\hline
Dim 2 & Motivates & [because, since]+ [he, she, they, I, you, we]+\\
An emotion or basic human drive that motivates X &&[feels, wants, likes] \\
\hline
Dim 3 & Enables & [because, since]+ [he, she, they, I, you, we]+ \\
A location state that enables X &&[is, was, were]+ [above, across from, \\
& & between, at, below, far from, in, in front of, \\
& & inside of,near, next to, on top of, outside of]\\
\hline
Dim 4 & Enables & [because, since]+[he, she, they, it, I, you, we]+ \\
A possession state that enables X & & [has, have]\\
\hline
Dim 5 & Enables & [because, since]+[he, she, they, it, I, you, we]+ \\
Other attribute that enables X &&[am, is, are, has, have, want, wants, need, needs] \\
\hline
Dim 6 & Causes/Enables & [causes, caused, results in , . This causes, . As a result] \\
An event that is directly caused or enabled by X && \\
\hline
Dim 7 & Causes & [. As a result]+ [he, she, they,I, you, we]+[feels] \\
An emotion that is caused by X && \\
\hline
Dim 8 & Results in & [. As a result]+ [he, she, they, it, I, you, we]+  \\ 
A change of location that X results in && between, [is, was, were]+ [above, across from,  \\
&& at, below, far from, in, in front of, \\
& & inside of,near, next to, on top of, outside of] \\
\hline
Dim 9 & Results in & [. As a result] + [he, she, they, it, I, you, we]+\\
A change of possession that X results in & & [has, have] \\
\hline
Dim 10 & Results in & [. As a result] + [he, she, they, it, I, you, we]+ \\
Other change in attribute that X results in & & [am, is, are, has, have, want, wants, need, needs]\\
\bottomrule
\end{tabular}
\end{center}
\caption{Templates used for turning the ten dimensions for GLUCOSE data into natural language statements for decoding proper sequences from the pre-trained language models. }\label{tab:decoding}
\end{table*}
\subsection*{1S-LM and Full-LM}
This model uses the exact model as with PT-LM. These models were finetuned on 8 NVIDIA Tesla V100 GPUs for 10K steps.  
\subsection*{Enc-Dec Model}
We finetuned the 770M-parameter pre-trained T5 model using the exact same hyperparameters as in \citep{2019t5}. We have used top-K random sampling for decoding, with K = 15. We did the training and decoding for this model on Google TPU v3-8. We trained this model for 500k steps after pre-training, which took about 72 hours.   

We also experimented with non-contextual version of all the models presented in the main paper. For non-contextual models, the story $S$ is simply removed from the input. The non-contextual models all underperformed their contextual counterparts. This further validates the importance of using context in making commonsense inferences.

\end{document}